\documentclass{article}
\usepackage{spconf,amsmath,graphicx}
\usepackage{amssymb}
\usepackage{pifont}
\usepackage{color}
\usepackage{booktabs}
\usepackage{multirow}
\usepackage{multicol}
\usepackage{lipsum}  
\usepackage{soul}

\usepackage[pagebackref,breaklinks,colorlinks]{hyperref}
\usepackage[capitalize]{cleveref}
\crefname{section}{Sec.}{Secs.}
\Crefname{section}{Section}{Sections}
\Crefname{table}{Table}{Tables}
\crefname{table}{Tab.}{Tabs.}

\title{ENHANCING MULTI-VIEW STEREO WITH CONTRASTIVE MATCHING AND WEIGHTED FOCAL LOSS}

\name{Yikang Ding\textsuperscript{\rm 1}\sthanks{Equal Contribution.}\ \ \ Zhenyang Li\textsuperscript{\rm 1}\footnotemark[1]\ \ \ Dihe Huang\textsuperscript{\rm 1}\footnotemark[1] \ \ Zhiheng Li\textsuperscript{\rm 1}\sthanks{Corresponding author (zhhli@tsinghua.edu.cn).} \ \ Kai Zhang\textsuperscript{\rm 1,}\textsuperscript{\rm 2}}
\address{\textsuperscript{\rm 1} Shenzhen International Graduate School, Tsinghua University\\
        \textsuperscript{\rm 2} Research Institute of Tsinghua, Pearl River Delta}

\begin{document}
%
\maketitle


\begin{abstract}
Learning-based multi-view stereo (MVS) methods have made impressive progress and surpassed traditional methods in recent years. However, their accuracy and completeness are still struggling. In this paper, we propose a new method to enhance the performance of existing networks inspired by contrastive learning and feature matching. First, we propose a \textbf{C}ontrast \textbf{M}atching \textbf{L}oss (CML), which treats the correct matching points in depth-dimension as positive sample and other points as negative samples, and computes the contrastive loss based on the similarity of features.
We further propose a \textbf{W}eighted \textbf{F}ocal \textbf{L}oss (WFL) for better classification capability, which weakens the contribution of low-confidence pixels in unimportant areas to the loss according to predicted confidence. Extensive experiments performed on DTU, Tanks and Temples and BlendedMVS datasets show our method achieves state-of-the-art performance and significant improvement over baseline network.
\end{abstract}
%
\section{Introduction}
\label{sec:intro}
Multi-view Stereo (MVS) aims to reconstruct the dense 3D presentation of the observed scene using a series of posed images and camera parameters, which plays an important role in augmented and virtual reality, robotics and 3D modeling. 
Though great progress has been made, these methods' performance is still struggling, especially in textureless regions. 
To solve this problem, some methods~\cite{wei2021aa,ding2022transmvsnet,liao2022wt-mvsnet} have made efforts to enhance the quality of the extracted features by leveraging deformable 2D CNNs and transformers. Although the performance of models can be improved, these additional modules increase the memory consumption and runtime. 

In this paper, we propose a new method to train the existing MVS networks inspired by contrastive learning and feature matching.
We analogize MVS back to its attribute of a feature matching task and propose to ameliorate the quality of extracted feature to enhance the feature matching in MVS.
Firstly, we propose Contrastive Matching Loss (CML) to enhance the feature matching quality. As shown in Fig. \ref{fig:overview-cml}, we apply CML when constructing cost volume from feature volume, which regards the correct candidate points in the depth dimension of source view as positive samples and other points in depth dimension as negative samples. By measuring the feature metric distance between the reference point and the positive and negative samples, we constrain the extracted feature of the reference point more similar to the positive sample, improving the quality of feature matching.
Secondly, we propose the Weighted Focal loss (WFL) to train the model for better classification capability. To do so, we transform the predicted confidence and take it as the weight of the original focal loss. Our insight is that the low-confidence (a.k.a high-uncertainty) pixels often appear in the background area or occlusion area, which usually brings in noise in the training phase. Reducing the influence of these high-uncertainty pixels helps the network pay close attention to the more important foreground pixels during training. As a result, our proposed CML and WFL significantly improve the performance of the baseline network. Extensive experiments performed on DTU~\cite{aanaes2016large}, Tanks and Temples~\cite{knapitsch2017tanks} and BlendedMVS~\cite{yao2020blendedmvs} datasets show our method achieves state-of-the-art performance.

\section{Related Work}
\label{sec:related-work}
\subsection{Learning based Multi-view Stereo}
With the development of deep learning, learning-based methods have been proposed to handle the task of MVS for better reconstruction quality. MVSNet~\cite{yao2018mvsnet} transforms the MVS task to a per-view depth map estimation task that encodes camera parameters via differentiable homography to build 3D cost volumes, which will be regularized by 3D CNN to obtain a probability volume and final depth. Inspired by MVSNet~\cite{yao2018mvsnet}, \cite{gu2020cascade,zhang2020visibility} follow this design paradigm. However, 3D U-Net architecture costs a lot of memory and runtime for cost volume regularization.
To solve this problem, several methods~\cite{yao2019recurrent,yan2020dense} proposed to replace 3D CNN with RNN and regularize the 3D cost volumes recurrently, adopting RNNs to pass features between different depth hypotheses.
\subsection{Contrastive learning}
Contrastive learning was first proposed in self-supervised learning. The main idea of contrastive learning is to maximize the consistency between pairs of positive samples and the difference between pairs of negative samples.
Contrastive learning becomes popular from Moco series\cite{chen2020moco}, which is proposed by Kaiming He. After that, SimCLR~\cite{chen2020SimCLR} also greatly improves the performance of classification network and self-supervised learning by transforming the input samples into positive samples and negative samples.

\begin{figure}[t]
  \centering
   \includegraphics[width=\linewidth]{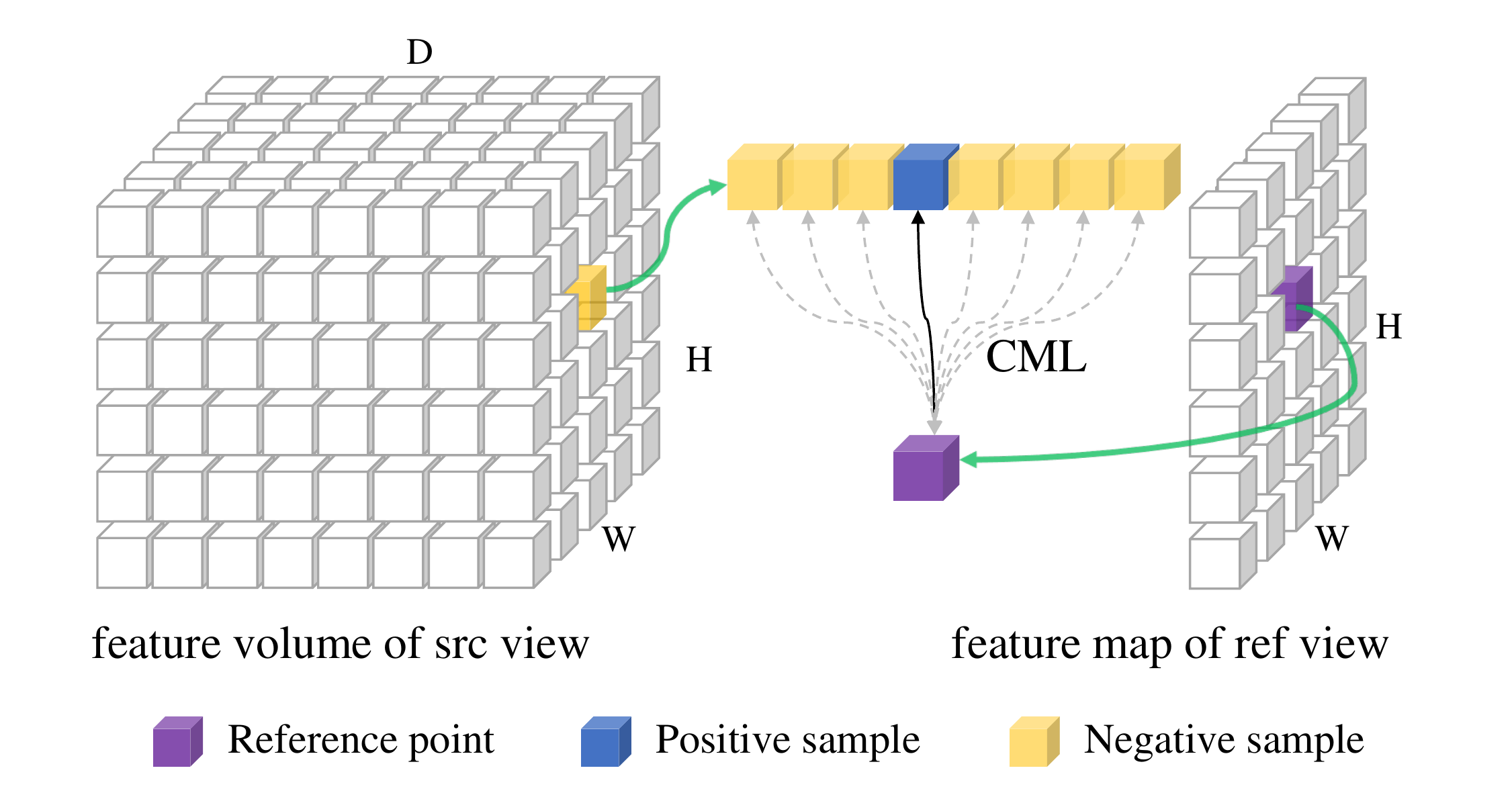}
   \caption{Visualization of Contrastive Matching Loss.}
   \label{fig:overview-cml}
\end{figure}

\section{Method}
\label{sec:method}

\subsection{Baseline Network}
\label{sec:network-overview}
Our method can be applied to arbitrary cost volume-based MVS network. We choose CasMVSNet~\cite{gu2020cascade} as the baseline network. Given the reference image $\mathbf{I}_0\in \mathbb{R}^{H\times W\times 3} $, the source images $\{\mathbf{I}_i\}_{i=1}^{N-1}$ neighboring it, and their camera intrinsics and extrinsics. CasMVSNet extracts the depth features of each image using FPN~\cite{lin2017feature} at first. Based on the candidate depth assumptions via homography warping, the baseline model warped the feature maps of $\{\mathbf{I}_i\}_{i=1}^{N-1}$ to the view of the reference image and get the feature volume. A variance-based metric will be used to aggregate all the feature volumes into one cost volume, and regularized by a 3D U-Net to get a probability volume. Finally, calculate the expectation of all the depth hypothesizes which from the probability volume to obtain the final depth.
More details of the baseline network could be found in~\cite{gu2020cascade}.

\subsection{Contrastive Matching Loss}
\label{sec:CML}
The key of cost volume-based MVS is obtaining the correct matching point among the candidate points. Therefore, we analogize MVS back to its attribute of a feature matching problem and propose to ameliorate the quality of extracted features to enhance the quality feature matching.
As shown in Fig. \ref{fig:overview-cml}, for each pixel $\mathbf{p}$ and its feature $f$ in reference view, we warp the feature map of the source view to the reference view via homography warping according to the candidate depth. The warping between a pixel $\mathbf{p}$ in reference view and its corresponding pixel $\hat{\mathbf{p}}_i$ in the $i^{th}$ source view under candidate depth $d$ in is defined as:
\begin{equation}
    \hat{\mathbf{p}}_{i} = \mathbf{K}_i[\mathbf{R_i}(\mathbf{K}_0^{-1}\mathbf{p}d)+\mathbf{t_i}],
\end{equation}
where $\mathbf{R}_i$ and $\mathbf{t}_i$ denote the rotation and translation between the $i^{th}$ source view and the reference view. $\mathbf{K}_0$ and $\mathbf{K}_i$ are the intrinsic matrices of the reference and the $i^{th}$ source camera.
Inspired by contrastive learning theory, we regard the feature of correct matching point $f^+$ as positive sample $x^+$ and the features of other points $\{f_{k}^-\}_{k=1}^{D-1}$ as negative samples $x^-$, where $D$ indicates the number of candidate depth. 

Based on the above description, we give the definition of our proposed CML as below:
\begin{equation}
		\mathcal{L}_{CML} = -\sum_{i=1}^{N-1}\left(sim(f^+_i,f)-\frac{1}{D-1}\sum_{k=1}^{D-1}sim(f^-_{i,k},f)\right),
\end{equation}
where $N$ indicates the total number of input views and $sim(\cdot, \cdot)$ is the similarity of two feature vectors. By default, we choose the inner product similarity as $sim(\cdot, \cdot)$:
\begin{equation}
		sim(a,b)=<a,b>,
\end{equation}
where $a\in \mathbb{R}^{1\times C}$, $b\in \mathbb{R}^{1\times C}$ and $C$ presents the channel number of features.

\subsection{Weighted Focal Loss}
\label{sec:WFL}
Based on our analysis that the key of MVS is improving the quality of feature matching, we try to improve the performance from the perspective of classification task. Specifically, 
The final outputs of cost volume-based MVS network~\cite{gu2020cascade}, a probability volume, represents the predicted probability on candidate depths, which means that MVS is also a classification task in essence. Therefore, different from the existing 3D CNN-based methods that use L1 loss, we propose an uncertainty-aware Weighted Focal Loss (WFL) to enhance the model's classification ability. 

\begin{figure}[h]
  \centering
  \vspace{-0.5cm} 
   \includegraphics[width=1.0\linewidth]{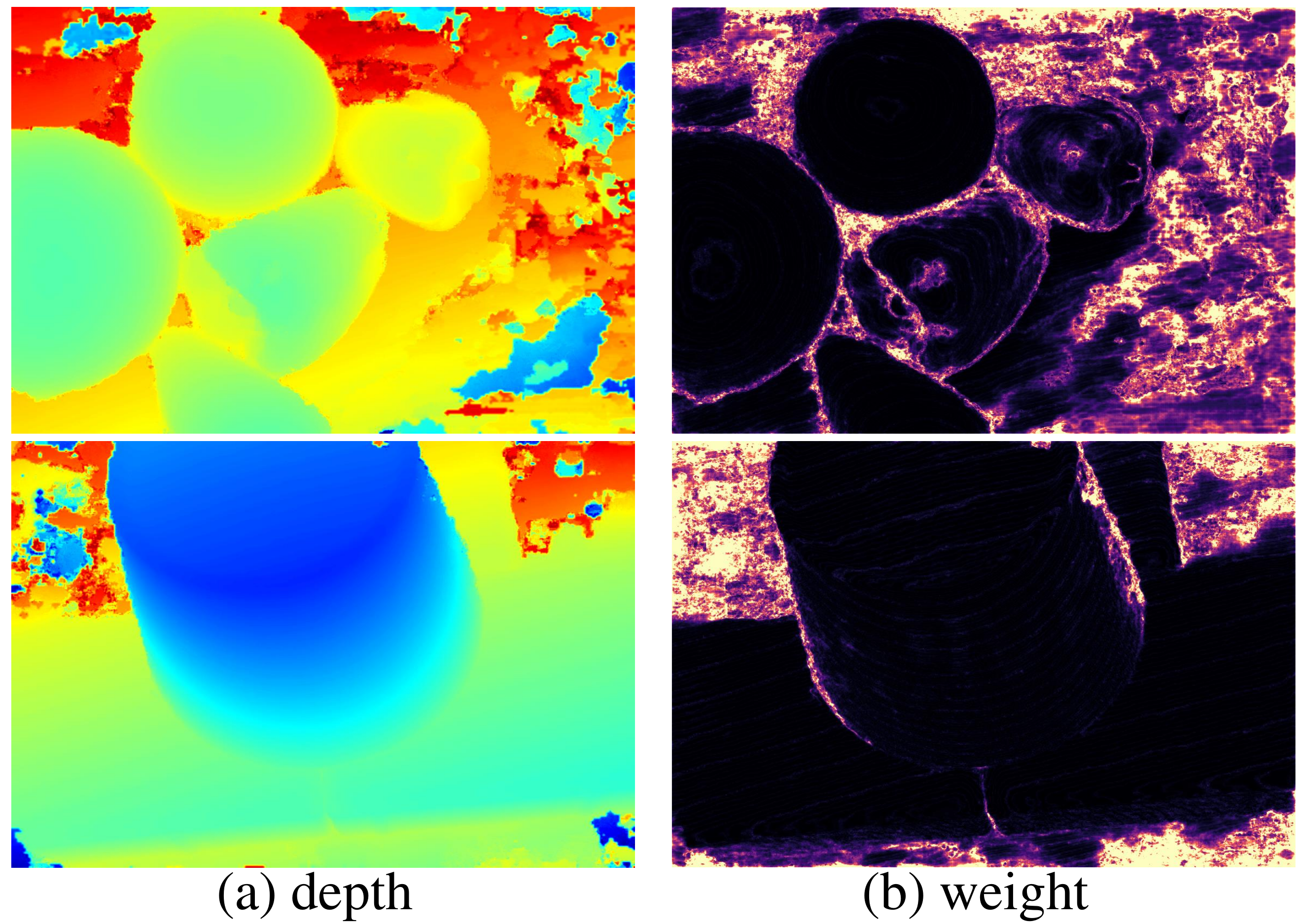}
   \caption{Visualization of weighted map in WFL.}
   \label{fig:weight-map}
\end{figure}

Our main insight is that the high-uncertainty (a.k.a low-confidence) pixels in unimportant regions (e.g. background) usually bring in noise during the training phase. To help the model pay close attention to the more important foreground regions, we transform the predicted confidence to weight different areas when compute focal loss:
\begin{equation}
\mathcal{L}_{WFL} = \left(\prod_{s=1}^3C_s^*\right)\sum_{\mathbf{p}\in \{\mathbf{p}_{v}\}}-(1-P^{(\tilde{d})}(\mathbf{p}))^2 \log\left(P^{(\tilde{d})}(\mathbf{p})\right),
\end{equation}
where $C_s^*$ indicated the upsampled confidence of $s$-th stage in CasMVSNet by bilinear interpolation. Multiplying each stage's confidence takes into account the confidential information of each stage and makes the final confidence more reliable. $P^{(d)}(\mathbf{p})$ denotes predicted probability of depth hypothesis $d$ at pixel $\mathbf{p}$ and $\tilde{d}$ represents the depth value closest to the ground truth among all hypotheses. $\{\mathbf{p}_{v}\}$ represents a subset of pixels with valid ground truth.
As shown in Fig. \ref{fig:weight-map}, our weight maps highlight the importance of the foreground region and suppress the influence of the background region, helping the model concentrate on the important foreground.

\begin{figure*}[t]
  \centering
   \includegraphics[width=\linewidth]{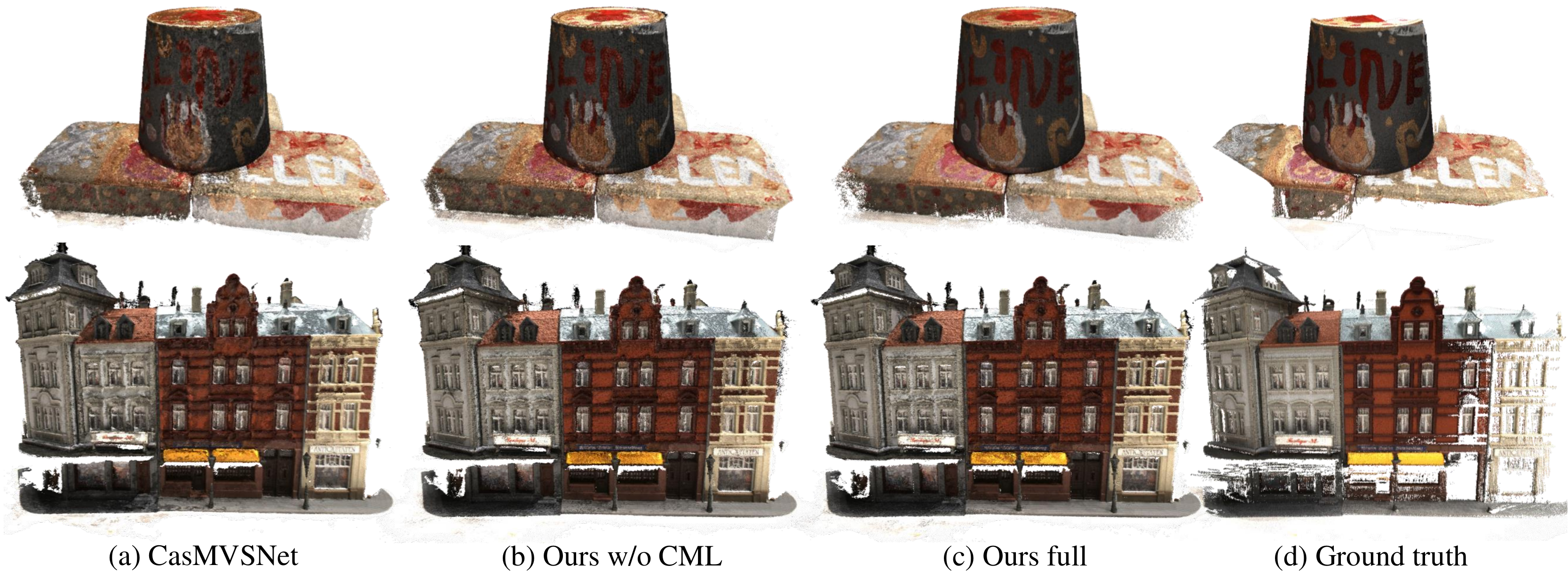}
   \vspace{-0.8cm} 
   \caption{Comparison of reconstructed results with baseline network~\cite{gu2020cascade} on DTU evaluation set~\cite{aanaes2016large}.}
   \label{fig:pcd_compare}
\end{figure*}

\vspace{-0.15cm}
\section{Experiments}
\label{sec:experiments}
\vspace{-0.15cm}
\subsection{Datasets}
This paper uses three popular datasets for training or testing.
DTU~\cite{aanaes2016large} is an indoor dataset that is captured under well-controlled laboratory conditions with a fixed camera trajectory. We split the dataset into 79 training scans, 18 validation scans, and 22 evaluation scans by following~\cite{yao2018mvsnet}.
BlendedMVS dataset~\cite{yao2020blendedmvs} is a large-scale dataset and contains various objects and scenes.
Tanks and Temples~\cite{knapitsch2017tanks} is a public benchmark acquired in realistic conditions, which contains 8 scenes for intermediate subset and 6 scenes for advanced subset.

\vspace{-0.25cm}
\subsection{Implementation Details}
We implement our method with PyTorch and choose CasMVSNet~\cite{gu2020cascade} as our baseline network. 
We train the network with Adam for 16 epochs with an initial learning rate of 0.001, which decays by a factor of 0.5 respectively after 8, 10, and 12 epochs as doing in~\cite{gu2020cascade}.

\begin{table}[t]
    \centering
    \resizebox{\linewidth}{!}{
    \begin{tabular}{l|c|c|c}
    \hline
    \textbf{Method} & Acc.($mm$) & Comp.($mm$) & Overall($mm$)\\
    \hline
    COLMAP~\cite{schonberger2016mvs} & 0.400 & 0.664 & 0.532 \\
    MVSNet~\cite{yao2018mvsnet} & 0.396 & 0.527 & 0.462 \\ 
    R-MVSNet~\cite{yao2019recurrent} & 0.385 & 0.459 & 0.422 \\ 
    $D^2$HC-RMVSNet~\cite{yan2020dense} & 0.395 & 0.378 & 0.386 \\
    PointMVSNet~\cite{chen2019point} & 0.342 & 0.411 & 0.376 \\ 
    Vis-MVSNet~\cite{zhang2020visibility} & 0.369  & 0.361 & 0.365 \\
    AA-RMVSNet~\cite{wei2021aa} & 0.376 & 0.339 & 0.357 \\
    CasMVSNet~\cite{gu2020cascade} & 0.325 &  0.385 & 0.355 \\
    EPP-MVSNet~\cite{ma2021epp} & 0.413 & \underline{0.296} & 0.355 \\
    PatchmatchNet~\cite{wang2021patchmatchnet} & 0.427 & \textbf{0.277} & 0.352 \\
    UCS-Net~\cite{cheng2020deep} & 0.338 & 0.349 & \underline{0.344} \\
    \hline
    \rule{0pt}{9pt}\
    \textbf{Ours+CasMVSNet} &  \underline{0.323}(\textcolor{red}{-0.003}) & 0.347(\textcolor{red}{-0.038}) & \textbf{0.335}(\textcolor{red}{-0.020}) \\[2pt]
   \hline
    \end{tabular}}
    \caption{Quantitative results on DTU evaluation set~\cite{aanaes2016large} (\textbf{lower is better}). \textbf{Bold} figures indicate the best, \underline{underlined} figures indicate the second best and \textcolor{red}{red} figures indicate our improvement over the baseline network~\cite{gu2020cascade}.}
    \label{tab:dtu}
\end{table}

\begin{table*}[t]
    \centering
    \setlength\tabcolsep{1.8pt}
    \resizebox{\linewidth}{!}{
    \begin{tabular}{l|c|cccccccc|c|cccccc}
    \hline
    \hline
    \textbf{Method} &  \textbf{Int.Mean} & Family & Francis & Horse & L.H. & M60 & Panther & P.G. & Train & \textbf{Adv.Mean} & Auditorium & Ballroom & Courtroom & Museum & Palace & Temple \\
    \hline
    COLMAP~\cite{schonberger2016mvs} & 42.14 & 50.41 & 22.25 & 26.63 & 56.43 & 44.83 & 46.97 & 48.53 & 42.04 & 27.24 & 16.02 & 25.23 & 34.70 & 41.51 & 18.05 & 27.94\\
    ACMM~\cite{xu2019multi}&57.27	&	69.24	&51.45	&46.97	&63.20&	55.07&	57.64&	60.08&	54.48 & 34.02 & 23.41&32.91& \underline{41.17} &48.13&23.87&34.60\\
    DeepC-MVS~\cite{kuhn2020deepc} &	59.79&	71.91&	54.08&	42.29&	 \textbf{66.54} &55.77& \textbf{67.47} &	60.47& \textbf{59.83} & 34.54 & \textbf{26.30} & 34.66 & \textbf{43.50} & 45.66 & 23.09 & 34.00\\
    AttMVS~\cite{luo2020attention} &	{60.05} &	73.90 &	 \underline{62.58} &	44.08 &	\underline{64.88} &	56.08&	59.39 &	 \textbf{63.42} & 56.06 & 31.93 & 15.96 &27.71& 37.99 & \textbf{52.01} & 29.07 & 28.84\\
    \hline
    R-MVSNet~\cite{yao2019recurrent} & 50.55 & 73.01 & 54.46 & 43.42 & 43.88 & 46.80 & 46.69 & 50.87 & 45.25 & 29.55 & 19.49 & 31.45 & 29.99 & 42.31 & 22.94 & 31.10\\
    PatchmatchNet~\cite{wang2021patchmatchnet} & 53.15 & 66.99 & 52.64 & 43.24 & 54.87 & 52.87 & 49.54 & 54.21 & 50.81 & 32.31 & 23.69 & 37.73 & 30.04 & 41.80 & 28.31 & 32.29\\
    CasMVSNet~\cite{gu2020cascade} & 56.84	&	76.37&	58.45&	46.26&	55.81&	56.11&	54.06&	58.18&	49.51 & 31.12 & 19.81 & 38.46 & 29.10 & 43.87 & 27.36 & 28.11\\
    Vis-MVSNet~\cite{zhang2020visibility}& 60.03 &	77.40&	60.23&	47.07&	63.44&	\underline{62.21} &	57.28&	{60.54}&	52.07 & 33.78 & 20.79 &	38.77 &	32.45 & 44.20 & 28.73 & 37.70\\
    AA-RMVSNet~\cite{wei2021aa} &	61.51 & 77.77 &	59.53&	\underline{51.53} &	64.02&	\textbf{64.05} & 59.47&	60.85&	54.90 & 33.53&20.96 & \underline{40.15} & 32.05 & 46.01 & 29.28 & 32.71\\
    EPP-MVSNet~\cite{ma2021epp} & \underline{61.68} & \underline{77.86} & 60.54 & \textbf{52.96} & 62.33 & 61.69 & \underline{60.34} & \underline{62.44} & 55.30 & \underline{35.72} & 21.28 & 39.74 & 35.34 & \underline{49.21} & \underline{30.00} & \textbf{38.75}\\
    \hline
    \rule{0pt}{12pt} 
    \textbf{Ours+CasMVSNet} & \textbf{61.75}(\textcolor{red}{+4.91}) & \textbf{79.51} & \textbf{64.15} & 50.82 & 62.19 & 62.20 & 60.04 & 56.87 & \underline{58.26} & \textbf{35.97}(\textcolor{red}{+4.85}) & \underline{25.16} & \textbf{42.78} & 34.62 & 46.54 & \textbf{30.86} & \underline{35.84}\\[3pt]
    \hline
    \hline
    \end{tabular}}
    \caption{Benchmarking results on the Tanks and Temples~\cite{knapitsch2017tanks}. The evaluation metric is mean F-score (\textbf{higher is better}). \textbf{Bold} figures indicate the best,  \underline{underlined} figures indicate the second best and \textcolor{red}{red} figures indicate improvement over ~\cite{gu2020cascade}.}
    \label{tab:tnt}
\end{table*}

\begin{table}[t]
    \centering
    \footnotesize
    \setlength{\tabcolsep}{2mm}
    \begin{tabular}{l|ccc}
    \hline
    \textbf{Method} & EPE & $e_1$ & $e_3$ \\
    \hline
    MVSNet~\cite{yao2018mvsnet} & 1.49 & 21.98 & 8.32 \\
    CVP-MVSNet~\cite{yang2020cost} & 1.90 &  19.73 & 10.24 \\
    CasMVSNet~\cite{gu2020cascade} & 1.43 &  19.01 & 9.77 \\ 
    Vis-MVSNet~\cite{zhang2020visibility} & 1.47  & 15.14 & 5.13 \\
    EPP-MVSNet~\cite{ma2021epp} & 1.17 & 12.66 & 6.20 \\
    \hline
    \rule{0pt}{9pt}
    \textbf{Ours+CasMVSNet} & \textbf{1.08}(\textcolor{red}{-0.35})  & \textbf{11.12}(\textcolor{red}{-7.89}) & \textbf{4.36}(\textcolor{red}{-5.41}) \\[2pt]
   \hline
    \end{tabular}
    \caption{Quantitative results towards predicted depth maps on BlendedMVS validation set~\cite{yao2020blendedmvs} (\textbf{lower is better}). \textcolor{red}{Red} figures indicate improvement over~\cite{gu2020cascade}.}
    \label{tab:bld}
    \vspace{-0.3cm} 
\end{table}

\vspace{-0.15cm} 
\subsection{Experimental results}
\subsubsection{Evaluation on DTU dataset}
We experiment our method on the evaluation partition of DTU dataset\cite{aanaes2016large} using the official metrics. We set $N = 5$ and the input resolution as $864 \times 1152$ at evaluation phase following~\cite{gu2020cascade}. Quantitative comparison is shown in \cref{tab:dtu}. Considering the point-cloud pairs from MVS reconstruction and ground truth, the mean absolute distance of all pairs is the method of Accuracy measures, while the Completeness measures is the opposite. At last, the overall performance of models is the average of Accuracy and Completeness measures. 
Compared with baseline network~\cite{gu2020cascade}, our method significantly improves the completeness and overall performance. The qualitative comparison of the reconstructed point cloud can be found in Fig. \ref{fig:pcd_compare}.

\vspace{-0.15cm} 
\subsubsection{Benchmarking on Tanks and Temples}
The experiment on Tanks and Temples benchmark~\cite{knapitsch2017tanks} is used to demonstrate the generalization ability of our method.
The camera parameters, depth ranges, and neighboring view selection are aligned with R-MVSNet~\cite{yao2019recurrent}. The original resolution of the images is required for the inference process. Quantitative comparisons on Tanks and Temples are shown in \cref{tab:tnt} and the metrics are mean F-score. Compared to baseline network and other learning-based works, our method achieves state-of-the-art performance on both intermediate set and advanced set.

\vspace{-0.15cm} 
\subsubsection{Evaluation on BlendedMVS dataset}
We demonstrate the quality of depth maps on BlendedMVS validation dataset~\cite{yao2020blendedmvs}. We set $N=5$ and image resolution as $512\times 640$, and apply the evaluation metrics described in \cite{darmon2021deep}. The quantitative results are illustrated in \cref{tab:bld}. The endpoint error is represented as EPE, which is the average $\ell$-1 distance of prediction and the ground truth depth; $e_1$ and $e_3$ represent the proportion in $\%$ of pixels with depth error larger than 1 and larger than 3. As shown in table \ref{tab:bld}, our method significantly improves the depth quality compared with baseline network and other coarse-to-fine methods.

\vspace{-0.35cm} 
\subsection{Ablation Study}
\vspace{-0.25cm} 
We provided the ablation studies to demonstrate the effectiveness of different components in our model. We trained our baseline model CasMVSNet~\cite{gu2020cascade} with $\ell$-1 loss, while all the experiments have the same hyperparameters. As shown in Tab. \ref{tab:ablation}, CML and WFL individually improved the overall performance over the baseline network (referred to as (a)). When CML and WFL are applied jointly, the overall performance is further significantly improved. 
\begin{table}[t]
    \centering
    \setlength\tabcolsep{2pt}
    \resizebox{\linewidth}{!}{
    \begin{tabular}{ccc|ccc|ccc}
    \hline
      \multicolumn{1}{}{aaa} & \multicolumn{2}{c|}{\textbf{Method Settings}}  & \multicolumn{3}{c|}{\textbf{DTU}} & \multicolumn{3}{c}{\textbf{BlendedMVS}}\\ 
      & CML & WFL & Acc.  & Comp. & Overall & EPE & $e_1$ & $e_3$ \\
       \hline
       (a) & & &  0.325 & 0.385 & 0.355 & 1.43 & 19.01 & 9.77 \\
       (b) & \checkmark &  & \textbf{0.323}  & 0.362  & 0.343  & 1.24  &  14.02  & 5.32  \\
       (c) & & \checkmark &  0.324  & 0.370  & 0.347  & 1.31  & 13.98  &  5.71 \\
       (d) & \checkmark & \checkmark& \textbf{0.323} & \textbf{0.347} & \textbf{0.335} & \textbf{1.08} & \textbf{11.12} & \textbf{4.36} \\
    \hline
    \end{tabular}}
    \caption{Ablation study on DTU evaluation dataset~\cite{aanaes2016large} and BlendeMVS evaluation dataset~\cite{yao2020blendedmvs}. (a) is trained with $\ell$-1 loss, (b) is with $\ell$-1 loss and CML, (c) is trained with only WFL and (d) is trained with CML and WFL.}
    \label{tab:ablation}
    \vspace{-0.5cm} 
\end{table}

\vspace{-0.4cm} 
\section{Conclusion}
\label{sec:conclusion}
\vspace{-0.25cm} 
In this paper, we propose a new method to train the existing MVS networks inspired by contrastive learning and feature matching.
The proposed CML ameliorates the distinguishability and robustness of the extracted features by constraining the reference feature more similar with to positive sample. 
We propose the WFL for better classification capability, which transforms the predicted confidence and takes it as the weight of the focal loss.
Extensive experiments show our method achieves state-of-the-art performance on multiple datasets.

\vspace{-0.25cm} 
\section{Acknowledgments}
\vspace{-0.25cm} 
This work is supported by Key-Area Research and Development Program of Guangdong (No.2020B0909050003) and Integration Project of Cognitive Computing of Audio Visual Information of Shenzhen (CJGJZD20200617102801005).

\bibliographystyle{IEEE}
\bibliography{egbib}

\end{document}